\documentclass[letterpaper]{article} 
\usepackage{aaai24}  
\usepackage{times}  
\usepackage{helvet}  
\usepackage{courier}  
\usepackage[hyphens]{url}  
\usepackage{graphicx} 
\usepackage{natbib}  
\usepackage{caption} 
\usepackage{algorithm}
\usepackage{algorithmic}
\usepackage{newfloat}
\usepackage{listings}
\DeclareCaptionStyle{ruled}{labelfont=normalfont,labelsep=colon,strut=off} 
\floatstyle{ruled}
\newfloat{listing}{tb}{lst}{}
\floatname{listing}{Listing}
\usepackage{booktabs}
\usepackage{multirow}
\usepackage{multicol}

\title{SecQA: A Concise Question-Answering Dataset\\for Evaluating Large Language Models in Computer Security}
\author {
    Zefang Liu
}
\affiliations{
    Georgia Institute of Technology\\
    North Avenue, Atlanta, GA 30332 USA\\
    liuzefang@gatech.edu\\
}

\begin{document}
\nocopyright
\maketitle
\begin{abstract}

In this paper, we introduce SecQA, a novel dataset tailored for evaluating the performance of Large Language Models (LLMs) in the domain of computer security. Utilizing multiple-choice questions generated by GPT-4 based on the ``Computer Systems Security: Planning for Success'' textbook, SecQA aims to assess LLMs' understanding and application of security principles. We detail the structure and intent of SecQA, which includes two versions of increasing complexity, to provide a concise evaluation across various difficulty levels. Additionally, we present an extensive evaluation of prominent LLMs, including GPT-3.5-Turbo, GPT-4, Llama-2, Vicuna, Mistral, and Zephyr models, using both 0-shot and 5-shot learning settings. Our results, encapsulated in the SecQA v1 and v2 datasets, highlight the varying capabilities and limitations of these models in the computer security context. This study not only offers insights into the current state of LLMs in understanding security-related content but also establishes SecQA as a benchmark for future advancements in this critical research area.

\end{abstract}

\section{Introduction}

As digital infrastructures become increasingly complex and integral to all aspects of modern society, the importance of robust computer security systems cannot be overstated. In this context, the advent of Large Language Models (LLMs) such as GPT-3.5 \cite{ouyang2022training}, GPT-4 \cite{openai2023gpt}, and their contemporaries presents a novel avenue for both potential vulnerabilities and solutions in security protocols. The ability of these models to understand, interpret, and predict human language has opened new frontiers in various domains, including computer security. However, evaluating these models' understanding and application of complex and nuanced security principles remains challenging.

To address this gap, we introduce SecQA\footnote{https://huggingface.co/datasets/zefang-liu/secqa}, a concise yet comprehensive dataset explicitly designed to evaluate the performance of LLMs in the domain of computer security. Drawing from the textbook ``Computer Systems Security: Planning for Success'' \cite{tolboom2023computer}, SecQA comprises a series of multiple-choice questions that have been generated and refined using GPT-4. This dataset is not just a testbed but a benchmarking tool that reflects the real-world complexities and scenarios encountered in computer security.

SecQA is structured into two distinct versions: v1 and v2. Version 1 offers a foundational level of assessment, focusing on the LLMs' basic understanding and application of security principles. In contrast, Version 2 escalates the challenge by presenting more complex and nuanced questions, thus pushing the models to demonstrate a deeper and more comprehensive understanding of advanced security topics.

In this paper, we detail the creation process of the SecQA dataset, its structure, and the rationale behind its design. We also present a comprehensive evaluation of several leading LLMs using SecQA under various settings, including 0-shot and 5-shot learning. This evaluation not only indicates the current capabilities of these models in the security domain but also serves as a baseline for future advancements.

Through our in-depth analysis and discussion, we aim to illuminate the strengths and weaknesses of current LLMs in understanding computer security and provide a path forward for researchers and developers in this critical field. By establishing SecQA as a standard benchmark, we encourage ongoing evaluation and improvement of LLMs, fostering a more secure and reliable digital future.

\section{Related Work}

The application of Large Language Models (LLMs) and Natural Language Processing (NLP) in cybersecurity \cite{kaddour2023challenges} is a dynamic area of research, transitioning from initially analyzing and understanding cybersecurity texts \cite{alam2022cyner,wang2022aptner} to encompassing broader tasks such as malware classification \cite{rahali2021malbert,rahali2023malbertv2}, threat detection \cite{ferrag2023revolutionizing}, vulnerability assessment \cite{shahid2021cvss,das2021v2w}, and command shell security \cite{andrew2022mapping,liu2023anomaly}. These models are increasingly recognized for their potential to interpret complex security-related data and make informed decisions, marking a significant evolution from their early use cases.

As Large Language Models (LLMs) grow more capable, the need for specialized datasets to benchmark their cybersecurity performance is more pronounced. While existing resources such as MMLU \cite{hendrycks2020measuring}, HELM \cite{liang2022holistic}, and BIG-bench\cite{srivastava2023beyond} address various language and domain-specific tasks, a notable gap remains for datasets that thoroughly evaluate LLMs' understanding of complex cybersecurity issues. Most current datasets focus on general language, overlooking the intricate spectrum of cybersecurity challenges. This highlights the necessity for a well-rounded, security-focused evaluation tool to assess and enhance LLMs' proficiency in the nuanced realm of cybersecurity.

Introduced against the backdrop of evolving LLM applications and the pressing need for specialized cybersecurity benchmarks, SecQA emerges as a novel dataset crafted to fill the void in comprehensive LLM evaluation. It provides a structured and challenging set of questions derived from authoritative security literature, aiming to rigorously assess LLMs' abilities to understand and apply complex computer security principles. This dataset not only bridges the critical gap between general language models and the specific demands of cybersecurity tasks but also represents a targeted initiative to enhance the application of LLMs in security-sensitive environments. By providing a nuanced benchmark for assessing LLMs in the context of computer security, SecQA guides the development of models adept at handling language intricacies and equipped to tackle the unique challenges presented by cybersecurity threats.

\section{Dataset Overview}

SecQA is designed as a specialized resource to critically evaluate the understanding and application of computer security principles by Large Language Models (LLMs). It aims to provide a concise benchmark that reflects the nuanced and complex nature of the security domain. This section outlines the dataset's composition, generation process, and structure, highlighting its potential as a tool for assessing LLMs.

\subsection{Composition and Generation}

SecQA is comprised of multiple-choice questions, each crafted to probe various aspects of computer security knowledge. The questions are generated using GPT-4 \cite{openai2023gpt}, leveraging its advanced language understanding and generation capabilities. To ensure the dataset's relevance and rigor, the questions are based on content extracted from the ``Computer Systems Security: Planning for Success'' textbook \cite{tolboom2023computer}. This approach ensures that the questions are grounded in established security principles and reflect real-world scenarios and challenges.

The generation process for the SecQA dataset employs a tiered approach, beginning with the creation of two specialized GPTs: Cyber Quizmaster for SecQA v1 and Cyber Quizmaster Pro for SecQA v2. The Cyber Quizmaster, developed using the GPT Builder\footnote{https://openai.com/blog/introducing-gpts} and informed by the ``Computer Systems Security: Planning for Success'' textbook, is tasked with producing a set of multiple-choice questions. It is meticulously programmed to ensure originality and adherence to learning objectives, generating questions that assess a foundational understanding of computer security principles. For the more advanced SecQA v2, the Cyber Quizmaster Pro takes over, engineered to challenge users further by generating more complex and nuanced questions, pushing the depth of understanding and application in the field.

Both versions receive specific learning objectives and then autonomously generate questions and answers based on the textbook content, structured for easy integration into the dataset. Each generated item is crafted to avoid direct textbook replication, balancing the difficulty and significance of each learning objective. Following their respective question generation phases, researchers conduct a thorough review, refinement, and validation process for both sets. This ensures that every question from both Cyber Quizmaster and Cyber Quizmaster Pro meets the standards of quality, relevance, and clarity. The result is a two-tiered dataset: SecQA v1 and v2, each tailored to different levels of complexity and understanding, providing a comprehensive tool for evaluating LLMs' reasoning, inference, and application skills in diverse security contexts.

\subsection{Dataset Structure}

SecQA is organized into two versions: v1 and v2. Version 1 is designed to assess foundational understanding, presenting questions that cover basic concepts and widely recognized security principles. It serves as a preliminary test to gauge LLMs' basic comprehension and application of security knowledge. Meanwhile, version 2 introduces a higher level of difficulty with more complex and nuanced questions, pushing LLMs to demonstrate a more profound understanding and advanced reasoning in this domain. This tiered approach allows researchers to assess LLMs across a spectrum of difficulty and provides insights into how models scale their understanding as complexity increases.

To facilitate a thorough evaluation process, SecQA is divided into development (dev), validation (val), and test splits for both versions. The Table \ref{tab:splits} below represents the distribution of questions across these splits. This segmentation ensures that models are tested against various scenarios and can robustly handle the breadth of challenges in computer security.

\begin{table}[h]
\centering
\begin{tabular}{lccc}
\toprule
\textbf{Version} & \textbf{Dev Set} & \textbf{Val Set} & \textbf{Test Set} \\
\midrule
SecQA v1 & 5 & 12 & 110 \\
SecQA v2 & 5 & 10 & 100 \\
\bottomrule
\end{tabular}
\caption{Distribution of questions in the development, validation, and test splits for SecQA v1 and v2.}
\label{tab:splits}
\end{table}

\subsection{Dataset Examples}

To provide a clear depiction of SecQA's content and its intended evaluative complexity, we present select questions from both SecQA v1 and SecQA v2 in the Table \ref{tab:examples}. These exemplars showcase the nuanced escalation in difficulty and conceptual depth achieved through the generation processes of Cyber Quizmaster and Cyber Quizmaster Pro.

\begin{table}[h]
\centering
\begin{tabular}{p{\linewidth}}
\toprule
\textbf{SecQA v1 Example} \\
\midrule
\textbf{Question:} What is the purpose of implementing a Guest Wireless Network in a corporate environment? \\
\textbf{A)} To provide unrestricted access to company resources \\
\textbf{B)} To offer a separate, secure network for visitors \\
\textbf{C)} To replace the primary corporate wireless network \\
\textbf{D)} To bypass network security protocols \\
\textbf{Correct Answer:} B \\
\textbf{Explanation:} A Guest Wireless Network provides visitors with internet access while segregating them from the main corporate network, enhancing security by preventing unauthorized access to sensitive company resources. \\
\midrule
\textbf{SecQA v2 Example} \\
\midrule
\textbf{Question:} What is a critical security consideration when implementing wireless security in an enterprise environment? \\
\textbf{A)} Ensuring the wireless network is completely isolated from the wired network. \\
\textbf{B)} Using the latest encryption standards, like WPA3, to protect wireless communication from eavesdropping and unauthorized access. \\
\textbf{C)} Prioritizing the range of the wireless network over its security. \\
\textbf{D)} Completely avoiding the use of wireless technology due to inherent security risks. \\
\textbf{Correct Answer:} B \\
\textbf{Explanation:} A critical aspect of wireless security is using robust encryption protocols to protect the data transmitted over wireless networks. Implementing the latest standards, such as WPA3, provides strong protection against eavesdropping and unauthorized access, ensuring the confidentiality and integrity of the wireless communication. \\
\bottomrule
\end{tabular}
\caption{Examples of questions from SecQA v1 and v2.}
\label{tab:examples}
\end{table}

\subsection{Potential and Applications}

SecQA acts as a stimulus for continuous innovation and enhancement in Large Language Models (LLMs). By providing a comprehensive evaluation of their performance in the nuanced domain of computer security, it illuminates both the strengths and areas for development within current models. This insight is vital for steering future advancements, ensuring that LLMs evolve in ways that are directly beneficial for tackling complex security tasks. While its primary role is to benchmark, SecQA's broader impact lies in its ability to guide the development of LLMs toward greater effectiveness and reliability, preparing them for an array of applications where understanding and responding to security-related content is crucial.

\section{Evaluation Results}

In this section, we conducted a preliminary evaluation of various Large Language Models (LLMs) on the SecQA v1 and v2 datasets to assess their ability to understand and apply computer security principles. The evaluation encompassed a range of LLMs, including GPT-3.5 \cite{ouyang2022training}, GPT-4 \cite{openai2023gpt}, Llama-2 \cite{touvron2023llama}, Vicuna \cite{zheng2023judging}, Mistral \cite{jiang2023mistral}, and Zephyr \cite{tunstall2023zephyr}, in both 0-shot and 5-shot learning settings. Accuracy served as the primary metric, providing a direct measure of each model's grasp of the concepts presented in the SecQA dataset.

The experiments, facilitated by the Language Model Evaluation Harness framework\footnote{https://github.com/EleutherAI/lm-evaluation-harness} \cite{gao2021framework}, were designed to understand the models' baseline capabilities without prior exposure (0-shot) and their adaptability when provided with limited context (5-shot). These scenarios are crucial for evaluating the potential real-world applications of LLMs, where they might encounter both familiar and unfamiliar security challenges. The results, summarized in the Table \ref{tab:results}, reflect the models' performance in these settings.

\begin{table}[h]
\centering
\begin{tabular}{lrr}
\toprule
\textbf{Model} & \textbf{0-Shot} & \textbf{5-Shot} \\
\midrule
\multicolumn{3}{c}{\textbf{SecQAv1 Dataset}} \\
\midrule
GPT-3.5-Turbo & 99.1 & 99.1 \\
GPT-4 & 99.1 & 100.0 \\
Llama-2-7B-Chat & 72.7 & 61.8 \\
Llama-2-13B-Chat & 49.1 & 89.1 \\
Vicuna-7B-v1.5 & 65.5 & 30.9 \\
Vicuna-13B-v1.5 & 76.4 & 40.0 \\
Mistral-7B-Instruct-v0.2 & 90.9 & 90.9 \\
Zephyr-7B-Beta & 84.6 & 92.7 \\
\midrule
\multicolumn{3}{c}{\textbf{SecQAv2 Dataset}} \\
\midrule
GPT-3.5-Turbo & 98.0 & 98.0 \\
GPT-4 & 98.0 & 98.0 \\
Llama-2-7B-Chat & 79.0 & 50.0 \\
Llama-2-13B-Chat & 51.0 & 89.0 \\
Vicuna-7B-v1.5 & 66.0 & 22.0 \\
Vicuna-13B-v1.5 & 74.0 & 42.0 \\
Mistral-7B-Instruct-v0.2 & 89.0 & 87.0 \\
Zephyr-7B-Beta & 81.0 & 86.0 \\
\bottomrule
\end{tabular}
\caption{Comparative accuracy of various Large Language Models (LLMs) on SecQA v1 and v2 datasets under 0-shot and 5-shot learning scenarios.}
\label{tab:results}
\end{table}

For SecQA v1, which focuses on fundamental security concepts, the results were generally positive, with certain models like GPT-4 achieving near-perfect accuracy in the 5-shot setting. This performance indicates a robust understanding of basic security principles and the ability to apply context effectively. However, the transition to the more challenging SecQA v2 revealed a broader range of performance across the models, with a notable decline in some cases. This shift highlights the complexity and nuanced understanding required for advanced security topics. For instance, while models like Llama-2-13B-Chat and Zephyr-7B-Beta showed promising improvement in the 5-shot setting for SecQA v2, others struggled, reflecting the variability in current LLMs' capabilities.

The evaluation results provide critical insights into the performance of current LLMs and the effectiveness of the SecQA dataset. While SecQA v1 confirms the dataset's ability to assess fundamental security knowledge, the near-perfect scores achieved by GPT-3.5-Turbo and GPT-4 on SecQA v2 raise important considerations about its challenge level, particularly since it was generated using GPT-4. This could limit its ability to rigorously evaluate GPT-4 due to potential inherent familiarity. Moreover, the varied performance, especially the struggles of some open-source LLMs on these specialized computer security questions, suggests the dataset's value in highlighting the necessity for domain-specific tailoring. This observation indicates that while SecQA offers valuable insights across complexities, it may need to provide more challenges for the advanced LLMs and highlight the need for domain-specific enhancements for others. To ensure ongoing relevance and rigor, SecQA will require continuous updates with more diverse and intricate scenarios. Such enhancements are essential to effectively test and benchmark LLMs, driving the development of models that are both proficient in general language tasks and finely attuned to the specificities of computer security.

\section{Conclusion}

This study introduces SecQA, a specialized dataset developed to assess the understanding of Large Language Models (LLMs) in the area of computer security. We tested a range of models, including GPT-3.5-Turbo, GPT-4, and open-source LLMs, and found a significant variation in their performance. While some models displayed a strong grasp of basic security concepts, they struggled with the more complex questions in SecQA v2. This highlights not only the progress LLMs have made in language understanding but also the room for growth in their ability to handle specialized, intricate knowledge.

The implications of our work are meaningful for both practical applications and future research. SecQA provides a concise benchmark for evaluating and improving LLMs in security-related tasks. The results emphasize the importance of ongoing development in targeted training and enhancing models' understanding of specific contexts. As the field continues to progress, the insights from this research will be critical for guiding the development of LLMs that are both more proficient and practical in addressing the complex challenges of computer security.

\bibliography{aaai24}

\begin{thebibliography}{21}
\providecommand{\natexlab}[1]{#1}

\bibitem[{Alam et~al.(2022)Alam, Bhusal, Park, and Rastogi}]{alam2022cyner}
Alam, M.~T.; Bhusal, D.; Park, Y.; and Rastogi, N. 2022.
\newblock CyNER: A Python Library for Cybersecurity Named Entity Recognition.
\newblock \emph{arXiv preprint arXiv:2204.05754}.

\bibitem[{Andrew, Lim, and Budiarto(2022)}]{andrew2022mapping}
Andrew, Y.; Lim, C.; and Budiarto, E. 2022.
\newblock Mapping Linux Shell Commands to MITRE ATT\&CK using NLP-Based
  Approach.
\newblock In \emph{2022 International Conference on Electrical Engineering and
  Informatics (ICELTICs)}, 37--42. IEEE.

\bibitem[{bench authors(2023)}]{srivastava2023beyond}
bench authors, B. 2023.
\newblock Beyond the Imitation Game: Quantifying and extrapolating the
  capabilities of language models.
\newblock \emph{Transactions on Machine Learning Research}.

\bibitem[{Das et~al.(2021)Das, Serra, Halappanavar, Pothen, and
  Al-Shaer}]{das2021v2w}
Das, S.~S.; Serra, E.; Halappanavar, M.; Pothen, A.; and Al-Shaer, E. 2021.
\newblock V2W-BERT: A Framework for Effective Hierarchical Multiclass
  Classification of Software Vulnerabilities.
\newblock In \emph{2021 IEEE 8th International Conference on Data Science and
  Advanced Analytics (DSAA)}, 1--12. IEEE.

\bibitem[{Ferrag et~al.(2023)Ferrag, Ndhlovu, Tihanyi, Cordeiro, Debbah, and
  Lestable}]{ferrag2023revolutionizing}
Ferrag, M.~A.; Ndhlovu, M.; Tihanyi, N.; Cordeiro, L.~C.; Debbah, M.; and
  Lestable, T. 2023.
\newblock Revolutionizing Cyber Threat Detection with Large Language Models.
\newblock \emph{arXiv preprint arXiv:2306.14263}.

\bibitem[{Gao et~al.(2021)Gao, Tow, Biderman, Black, DiPofi, Foster, Golding,
  Hsu, McDonell, Muennighoff et~al.}]{gao2021framework}
Gao, L.; Tow, J.; Biderman, S.; Black, S.; DiPofi, A.; Foster, C.; Golding, L.;
  Hsu, J.; McDonell, K.; Muennighoff, N.; et~al. 2021.
\newblock A Framework for Few-Shot Language Model Evaluation.
\newblock \emph{Version v0. 0.1. Sept}.

\bibitem[{Hendrycks et~al.(2020)Hendrycks, Burns, Basart, Zou, Mazeika, Song,
  and Steinhardt}]{hendrycks2020measuring}
Hendrycks, D.; Burns, C.; Basart, S.; Zou, A.; Mazeika, M.; Song, D.; and
  Steinhardt, J. 2020.
\newblock Measuring Massive Multitask Language Understanding.
\newblock \emph{arXiv preprint arXiv:2009.03300}.

\bibitem[{Jiang et~al.(2023)Jiang, Sablayrolles, Mensch, Bamford, Chaplot,
  Casas, Bressand, Lengyel, Lample, Saulnier et~al.}]{jiang2023mistral}
Jiang, A.~Q.; Sablayrolles, A.; Mensch, A.; Bamford, C.; Chaplot, D.~S.; Casas,
  D. d.~l.; Bressand, F.; Lengyel, G.; Lample, G.; Saulnier, L.; et~al. 2023.
\newblock Mistral 7B.
\newblock \emph{arXiv preprint arXiv:2310.06825}.

\bibitem[{Kaddour et~al.(2023)Kaddour, Harris, Mozes, Bradley, Raileanu, and
  McHardy}]{kaddour2023challenges}
Kaddour, J.; Harris, J.; Mozes, M.; Bradley, H.; Raileanu, R.; and McHardy, R.
  2023.
\newblock Challenges and Applications of Large Language Models.
\newblock \emph{arXiv preprint arXiv:2307.10169}.

\bibitem[{Liang et~al.(2022)Liang, Bommasani, Lee, Tsipras, Soylu, Yasunaga,
  Zhang, Narayanan, Wu, Kumar et~al.}]{liang2022holistic}
Liang, P.; Bommasani, R.; Lee, T.; Tsipras, D.; Soylu, D.; Yasunaga, M.; Zhang,
  Y.; Narayanan, D.; Wu, Y.; Kumar, A.; et~al. 2022.
\newblock Holistic Evaluation of Language Models.
\newblock \emph{arXiv preprint arXiv:2211.09110}.

\bibitem[{Liu and Buford(2023)}]{liu2023anomaly}
Liu, Z.; and Buford, J. 2023.
\newblock Anomaly Detection of Command Shell Sessions based on DistilBERT:
  Unsupervised and Supervised Approaches.
\newblock \emph{arXiv preprint arXiv:2310.13247}.

\bibitem[{OpenAI(2023)}]{openai2023gpt}
OpenAI. 2023.
\newblock GPT-4 Technical Report.
\newblock \emph{arXiv preprint arXiv:2303.08774}.

\bibitem[{Ouyang et~al.(2022)Ouyang, Wu, Jiang, Almeida, Wainwright, Mishkin,
  Zhang, Agarwal, Slama, Ray et~al.}]{ouyang2022training}
Ouyang, L.; Wu, J.; Jiang, X.; Almeida, D.; Wainwright, C.; Mishkin, P.; Zhang,
  C.; Agarwal, S.; Slama, K.; Ray, A.; et~al. 2022.
\newblock Training Language Models to Follow Instructions with Human Feedback.
\newblock \emph{Advances in Neural Information Processing Systems}, 35:
  27730--27744.

\bibitem[{Rahali and Akhloufi(2021)}]{rahali2021malbert}
Rahali, A.; and Akhloufi, M.~A. 2021.
\newblock MalBERT: Using Transformers for Cybersecurity and Malicious Software
  Detection.
\newblock \emph{arXiv preprint arXiv:2103.03806}.

\bibitem[{Rahali and Akhloufi(2023)}]{rahali2023malbertv2}
Rahali, A.; and Akhloufi, M.~A. 2023.
\newblock MalBERTv2: Code Aware BERT-Based Model for Malware Identification.
\newblock \emph{Big Data and Cognitive Computing}, 7(2): 60.

\bibitem[{Shahid and Debar(2021)}]{shahid2021cvss}
Shahid, M.~R.; and Debar, H. 2021.
\newblock CVSS-BERT: Explainable Natural Language Processing to Determine the
  Severity of a Computer Security Vulnerability from its Description.
\newblock In \emph{2021 20th IEEE International Conference on Machine Learning
  and Applications (ICMLA)}, 1600--1607. IEEE.

\bibitem[{Tolboom(2023)}]{tolboom2023computer}
Tolboom, R. 2023.
\newblock Computer Systems Security.

\bibitem[{Touvron et~al.(2023)Touvron, Martin, Stone, Albert, Almahairi,
  Babaei, Bashlykov, Batra, Bhargava, Bhosale et~al.}]{touvron2023llama}
Touvron, H.; Martin, L.; Stone, K.; Albert, P.; Almahairi, A.; Babaei, Y.;
  Bashlykov, N.; Batra, S.; Bhargava, P.; Bhosale, S.; et~al. 2023.
\newblock Llama 2: Open Foundation and Fine-Tuned Chat Models.
\newblock \emph{arXiv preprint arXiv:2307.09288}.

\bibitem[{Tunstall et~al.(2023)Tunstall, Beeching, Lambert, Rajani, Rasul,
  Belkada, Huang, von Werra, Fourrier, Habib et~al.}]{tunstall2023zephyr}
Tunstall, L.; Beeching, E.; Lambert, N.; Rajani, N.; Rasul, K.; Belkada, Y.;
  Huang, S.; von Werra, L.; Fourrier, C.; Habib, N.; et~al. 2023.
\newblock Zephyr: Direct Distillation of LM Alignment.
\newblock \emph{arXiv preprint arXiv:2310.16944}.

\bibitem[{Wang et~al.(2022)Wang, He, Xiong, Wei, Jiang, Chen, and
  Jiang}]{wang2022aptner}
Wang, X.; He, S.; Xiong, Z.; Wei, X.; Jiang, Z.; Chen, S.; and Jiang, J. 2022.
\newblock APTNER: A Specific Dataset for NER Missions in Cyber Threat
  Intelligence Field.
\newblock In \emph{2022 IEEE 25th International Conference on Computer
  Supported Cooperative Work in Design (CSCWD)}, 1233--1238. IEEE.

\bibitem[{Zheng et~al.(2023)Zheng, Chiang, Sheng, Zhuang, Wu, Zhuang, Lin, Li,
  Li, Xing, Zhang, Gonzalez, and Stoica}]{zheng2023judging}
Zheng, L.; Chiang, W.-L.; Sheng, Y.; Zhuang, S.; Wu, Z.; Zhuang, Y.; Lin, Z.;
  Li, Z.; Li, D.; Xing, E.~P.; Zhang, H.; Gonzalez, J.~E.; and Stoica, I. 2023.
\newblock Judging LLM-as-a-judge with MT-Bench and Chatbot Arena.
\newblock arXiv:2306.05685.

\end{thebibliography}
\end{document}